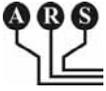

# Neural Networks in Mobile Robot Motion

**Danica Janglová**
Institute of Informatics SAS, danica.janglova@savba.sk

**Abstract:** This paper deals with a path planning and intelligent control of an autonomous robot which should move safely in partially structured environment. This environment may involve any number of obstacles of arbitrary shape and size; some of them are allowed to move. We describe our approach to solving the motion-planning problem in mobile robot control using neural networks-based technique. Our method of the construction of a collision-free path for moving robot among obstacles is based on two neural networks. The first neural network is used to determine the "free" space using ultrasound range finder data. The second neural network "finds" a safe direction for the next robot section of the path in the workspace while avoiding the nearest obstacles. Simulation examples of generated path with proposed techniques will be presented.

## 1. Introduction

Over the last few years, a number of studies were reported concerning a machine learning, and how it has been applied to help mobile robots to improve their operational capabilities. One of the most important issues in the design and development of intelligent mobile system is the navigation problem. This consists of the ability of a mobile robot to plan and execute collision-free motions within its environment. However, this environment may be imprecise, vast, dynamical and either partially or non-structured. Robots must be able to understand the structure of this environment. To reach their targets without collisions, the robots must be endowed with perception, data processing, recognition, learning, reasoning, interpreting, decision-making and action capacities. The ability to acquire these faculties to treat and transmit knowledge constitutes the key of a certain kind of artificial intelligence. Reproduce this kind of intelligence is, up to now, a human ambition in the construction and development of intelligent machines, and particularly autonomous mobile robots. To reach a reasonable degree of autonomy two basic requirements are sensing and reasoning. The former is provided by on-board sensory system that gathers information about robot with respect to the surrounding scene. The later is accomplished by devising algorithms that exploit this information in order to generate appropriate commands for the robot. And with this algorithm we will deal in this paper.
We report on the objective of the motion planning problem well known in robotics. Given an object with an initial location and orientation, a goal location and orientation, and a set of obstacles located in workspace, the problem is to find a continuous path from the initial position to the goal position, which avoids collisions with obstacles along the way. In other words, the motion planning problem is divided into two sub-problems, called 'Findspace' and 'Findpath' problem. For related approaches to the motion planning problem see reference (Latombe, J.C. 1991). The findspace problem is construction the configuration space of a given object and some obstacles. The findpath problem is in determining a collision-free path from a given start location to a goal location for a robot. Various methods for representing the configuration space have been proposed to solve the findpath problem (Brady, M. & all 1982), (Latombe, J.C. 1991), (Vörös, J. 2002). The major difficulties in the configuration space approach are: expensive computation is required to create the configuration space from the robot shape and the obstacles and the number of searching steps increases exponentially with the number of nodes. Thus, there is a motivation to investigate the use of parallel algorithms for solving these problems, which has the potential for much increased speed of calculations. A neural network is a massive system of parallel distributed processing elements connected in a graph topology. Several researchers have tried to use neural networks techniques for solving the find
path problem (Bekey, G.A. & Goldberg, K.Y., 1993).
In this paper we introduce a neural networks-based approach for planning collision-free paths among known stationary obstacles in structured environment for a robot with translational and rotational motion. Our approach basically consists of two neural networks to solve the findspace and findpath problems respectively. The first neural network is a modified principal component analysis network, which is used to determine the "free space" from ultrasound range finder data. Moving robot is modeled as a two-dimensional object in this workspace. The second one is a multilayer perceptron, which is used to find a safe direction for the next robot step on the collision-free path in the workspace from start configuration to a goal configuration while avoiding the obstacles.

The organization of the paper is as follows: section 2 brings out the briefly description of neural network applications in robotics. Our approach to solving the robot motion problem is given in section 3. Our method of motion planning strategy, which depends in using two neural networks for solving the findspace problem and the findpath problem respectively will be described in section 4. Simulation results will be included in section 5. Section 6 will summarize our conclusions and gives the notes for our further research in this area.

## 2. Neural networks in robotics

The interest in neural network stems from the wish of understanding principles leading in some manner to the comprehension of the basic human brain functions, and to building the machines that are able to perform complex tasks. Essentially, neural network deal with cognitive tasks such as learning, adaptation, generalization and optimization. Indeed, recognition, learning, decision-making and action constitute the principal navigation problems. To solve these problems fuzzy logic and neural networks are used. They improve the learning and adaptation capabilities related to variations in the environment where information is qualitative, inaccurate, uncertain or incomplete. The processing of imprecise or noisy data by the neural networks is more efficient than classical techniques because neural networks are highly tolerant to noises.

A neural network is a massive system of parallel distributed processing elements (neurons) connected in a graph topology. Learning in the neural network can be supervised or unsupervised. Supervised learning uses classified pattern information, while unsupervised learning uses only minimum information without preclassification. Unsupervised learning algorithms offer less computational complexity and less accuracy than supervised learning algorithms. Then former learn rapidly, often on a single pass of noisy data. The neural network could express the knowledge implicitly in the weights, after learning. A mathematical expression of a widely accepted approximation of the Hebbian learning rule is

$$w_{ij}(t+1) = w_{ij}(t) + \eta \, x_i(t) \, y_j(t) \qquad (1)$$

where $x_i$ and $y_j$ are the output values of neurons i and j, respectively, which are connected by the synapse $w_{ij}$ and

$\eta$ is the learning rate (note that $x_i$ is the input to the synapse).

Survey of types, architectures, basic algorithms and problems that may be solved using some type of neural networks is presented in (Jain, A.K. & Mao, J. & Mohiuddin, K.M. 1996). The applications of neural networks for classification and pattern recognition are good known. Some interesting solutions to problems of classification in the robot navigation domain were succesfully solved by means of competitive type of neural networks (Bekey, G.A. & Goldberg, K. Y. 1993). Using of competitive neural networks in control and trajectory generation for robots we may find in the book as well as using of neural network for sensor data processing in map updating and learning of the robot trajectories. For the obstacle avoidance purposes recurrent type of neural network was used with the gradient back-propagation technique for training the network (Domany, E. & Hemmen, J.L. & Schulten, K. 1991). The using of supervised neural network for robot navigation in partially known environment is presented in (Chochra 1997). An interesant solution with using of Jordan architecture of neural network is described in (Tani, J. 1996). Here the robot learns internal model of the environment by recurrent neural network, it predicts succession of sensors inputs and on the base of the model it generates navigation steps as a motor commands. The solution of the minimum path problem with two recurent neural networks is given in (Wang 1998). Solutions that use the learning ability of the neural network with fuzzy logic for representation of the human knowledge applied to robot navigation also exists see (Kim 1998). The complex view for solution of the navigation problem of the autonomous vehicles gives (Hebert, Thorpe, Ch. & Stentz, A. 1997). Team of researches CMU here presents results from designing of autonomous terrain vehicle. For learning the path from vision system data and for obstacle avoidances algorithms using laser range finder data and different types of neural networks.

Our first work concerned the using neural networks for object classification in the map of the robot environment was using the cluster analysis with range finder data (Uher, L. & Považan I. 1998). This acquiring knowledge we extend for using neural network in the algorithm of the robot motion planning.

## 3. The proposed approach

### 3.1. The basic motion planning problem

Let A be a rigid object, a robot, moving in a workspace W, represented as a subspace of $R^N$, with N=2 or 3. Let $O_1, ..., O_m$ be fixed rigid objects distributed in W, called obstacles. Assume that both the geometry and the location of A, $O_1, ..., O_m$, in W is known.

The problem is: Given an initial position and orientation of A in W, generate a path specifying a contiguous sequence of positions and orientations of A avoiding collision with $O_i$'s, starting at the initial position and orientation, and terminating at the goal position and orientation. Report the failure if no such path exists.

### 3.2. Environment representation

In general, we consider the case when A is a two-dimensional object that translates and rotates in $W=R^2$. A grid map will represent the environment. The grid map is an M x N matrix with each element representing the status $S_{i,j}$ of an individual grid; $S_{i,j} = 1$, if its interior intersects with the obstacle region and $S_{i,j} = 0$, if its interior does not intersect the obstacle region.

A configuration of an arbitrary object is a specification of the position of every point in this object relative to a fixed reference frame. In addition, let $F_A$ and $F_W$ be Cartesian frame embedded in A and W, respectively. Therefore, a configuration of A is a specification of the position (x,y) and orientation $\theta$ of $F_A$ with respect to $F_W$. Throughout the paper we make use of the localization system (Považan, I.; Janglova, D. & Uher, L. 1995) providing the robot with its absolute position with respect to a fixed inertial frame. The configuration space of A is the space of all the configurations of A. Let the resolution in x-axis, y-axis and orientation is M, N, and K respectively. A rectangloid $r_{i,j,k}$ is model of the object A located by $(x_i, y_j, \theta_k)$ and it represents the region $[x_i - w_x/2, x_i + w_x/2]$ . $[y_j - w_y/2, y_j + w_y/2]$ . $[\theta_k - \Delta\theta/2, [\theta_k + \Delta\theta/2]$, where $w_x$ is the width, $w_y$ is the height and $\Delta\theta = \pi/K$.

### 3.3. Motion planning algorithm

Philosophy of our algorithm appear from motion of human in the environment when he is moving between obstacles on the base of his eyes view and he make already the next step to the goal in the free space. Analogically, our robot will move safely in environment on the base of the data "visible" with scanning ultrasound range finder (Uher, L. & Kello, I. 1999). First must "mapping" the workspace from measured data and find the free space for robot motion and then determines the next robot azimuth for the safe step to the goal. For the solution of this problems we use neural networks technique. We use the measured range finder data in the learning workspace for mapping the front robot workspace by the first neural network finding the free space segment. This segment is used as an input to the second neural network both with the goal location, which is used to determine the direction of the proposed next navigation step for moving the robot.

The algorithm is of an iterative type. In each iteration, the last orientation of the moving robot is stored and the neural network selects the direction of the next navigation step. To determine the direction, the status in the partial configuration space is required; the map from range finder is proposed to give this status. Moreover, a control unit is used to provide information required by neural networks to control the operating sequence and to check the reachability of the goal configuration from the start configuration. Our motion planning algorithm can be summarized as follows:

1. Specify the object, environment information and the start and goal configurations.

2. Set the current object orientation equal to the goal orientation.

3. Activate range finder via control unit to determine the local part of the map of the workspace.

4. Initialize the first neural network, which will use the measured data from range finder. The neural network is iterated until the weights and the outputs converged to the returned one free space segment.

5. Activate the second neural network. It returns the direction $\theta_k$ of next robot motion step.

6. Generate the robot motion path in the direction $\theta_k$ and go to the step 3.

## 4. Principles of proposed algorithm

### 4.1. The findspace problem using neural network

Therefore we use the sensor data from the environment and the classical findspace problem in our strategy was transform to the procedure 'learning your environment'. The robot has in any position in workspace information about its distances to the all objects in this workspace. We use this information in first neural network that learns these situations and in any position gives the free segment of space for safe path as output. The neural network using for the findspace problem is principal component analysis network (PCA).

Principal component analysis networks combine unsupervised and supervised learning in the same topology (see Fig. 1).

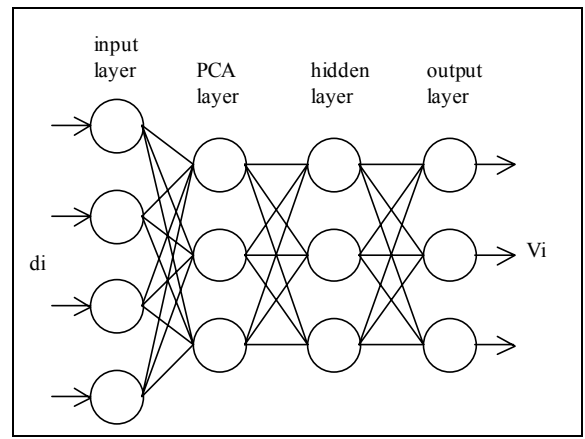

Fig. 1. PCA neural network topology
This neural network uses as inputs the data measured by the range finder. The output is free segment of the robot workspace.

Principal component analysis is an unsupervised linear procedure that finds a set of uncorrelated features from the input. A feed-forward network is used to perform the nonlinear classification from these components. PCA is a data reduction method, which condenses the input data down to a few principal components.

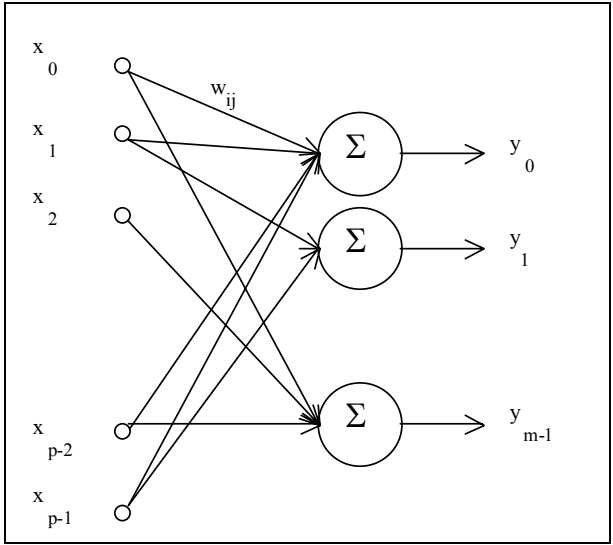

Fig. 2. Detail in PCA network

The number of principal components selected will be a compromise between training efficiency and accurate results. It is not possible to provide a general formula for selecting an appropriate number of principal components for a given application. Both learning rule choices are normalized implementations of the Hebbian learning rule. Straight Hebbian learning must be utilized with care, since it may become unstable if the inputs are not properly normalized. The network has four layers - input, PCA, hidden and output layer. The learning is realized in two phases.

In the first place an unsupervised linear procedure gets a set of uncorrelated features from the inputs and selects a few principal components. These components in hidden layer feed-forward supervised part gives the output. The PCA neural network learns by generalized Hebbian rule. First updated the synapse weights $w_{ij}$ (see Fig. 2) to a small random number and learning parameter $\eta$ to a positive small number. Then for n=1 are calculating outputs $y_j$ and the changes of the weights.

The outputs $y_j$ from PCA network are given by (2)

$$y_j(n) = \sum_{i=0}^{p-1} w_{ij}(n) x_i(n) \qquad j = 0,1,\ldots,m-1 \quad (2)$$

The changes of the weights during the learning are calculated by modification of Hebbian rule (3)

$$\Delta w_{ji}(n) = \eta \left[ y_j(n) x_i(n) - y_j(n) \sum_{k=0}^{j} w_{ki}(n) \, y_k(n) \right] \quad (3)$$

$$i = 0,1,\ldots,p-1$$

$$j = 0,1,\ldots,m-1$$

The calculations iterate up to the weights $w_{ij}$ are stable. At second phase the learning of the network are realised by back-propagation algorithm. Here updated the synaptic weights and treshold neuron coefficients. The back-propagation learning algorithm is based on the error-correction principle, i.e. it is necessary to know the network response to the input pattern. The learning process is as follows: On the input are given the input data and then are calculating the response of the network (feed-forward calculation). The error $e_i$ between an actual and an desired output is acquiring by formula (4)

$$e_i(n) = d_i(n) - y_i(n) \qquad (4)$$

where $d_i$ is desired output and $y_i$ is the actual output. During the back-propagation are compute the local gradient $\delta_i$ for the preceeding layers by propagating the errors backwards. Update the weights using formula (5)

$$w_{ij}(n+1) = w_{ij}(n) + \eta \delta_i(n) x_j(n) \quad (5)$$

where $w_{ij}$ is the weight between ith neuron from last layer and jth neuron of the next layer, $\eta$ is the learning rate. This process is repeated for the next input-ouput pattern until the error in the output layer is below a prespecified threshold or a maximum number of iterations are reached. We used the minimization of the average squared error cost function $E_{avg}$ given by (6)

$$E_{avg} = \frac{1}{N} \sum_{n=1}^{N} \frac{1}{2} \sum_{j=1}^{V} (d_j(n) - y_j(n))^2 \quad (6)$$

where N is number of the input-output patterns and V is the number of output neurons.

The neural network in that case uses the normalized data from ultrasound range finder as inputs. There are distances $d_i$, ranging from 20 to 250 cm, to the all objects in the front robot space from 0° to 180°. From input layer of the network we obtained information about free segments $V_i$. Each of the output neurons "represent" particular segment of the workspace as is depicted on the Fig. 3.

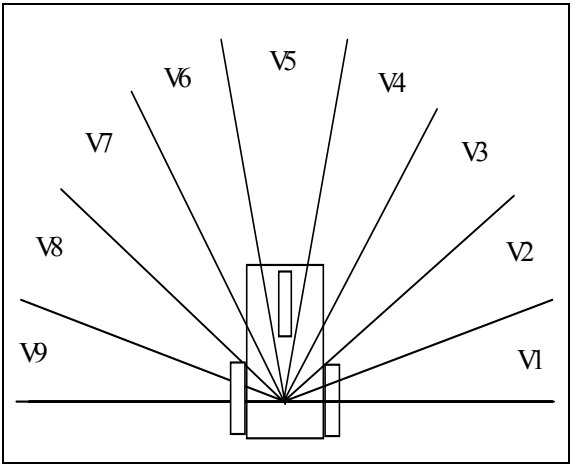

Fig. 3. The workspace segments

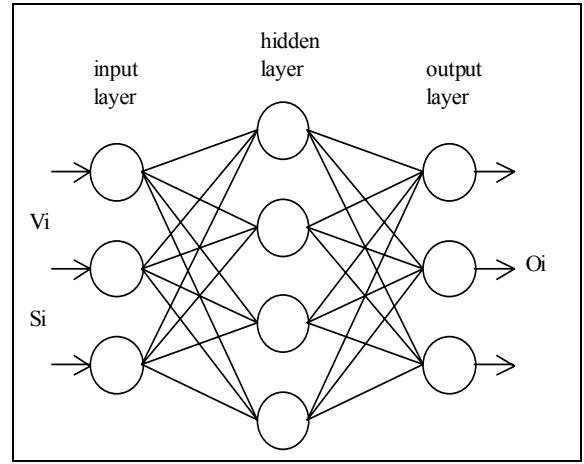

Fig. 4. Topology of MLP network

## 4.2. Solving the findpath problem

For the solving of the findpath problem we use neural network, too. Here we used as a neural network a multilayer perceptron (MLP). Multilayer perceptrons are layered feed-forward networks typically trained with static back-propagation. These networks have found their way into countless applications requiring static pattern classification. Their main advantage is that they are easy to use, and that they can approximate any input/output map. The key disadvantages are that they train slowly, and require lots of training data.

Aim of this network is determining of the robot azimuth $\theta_k$ for the next robot motion step from the output of first network and from the goal coordinates. The topology of this network is depicted on the Fig. 4. The network contains a three layer – input, hidden and output. This is a layered feed-forward network typically trained with static back-propagation. Here are updating the synapse weights between neurons and threshold. The process is similar to above described for the second phase of the learning process for findspace problem.

On the input of this neural network in our case we give the known free space segment $V_i$ as the output of the first neural network and the goal segments $S_i$ in which the coordinates of the robot goal position should be situated. The choice of the goal segments is the same as is depicted on the Fig. 3. From output layer of this neural network we obtained information $O_j$ about robot motion direction (azimuth) in the next step.

This information is given to the control unit. It manages this information into the robot command for the robot motor control.

## 4.3. Realization of the proposed algorithm

The proposed algorithm was prepared as a program Neuro in Microsoft Visual C++ with operating system Windows. For the learning and testing of the network we used the programs from NeuroSolutions packages (Neurodimension 2000). The program Neuro will secure the workspace visualization and robot motion simulation.

The learning of all neural networks was realized off-line with scanning data in the work environment. We use a more type of the learning environment. As a basic environment for "learning your environment" was used environment that is depicted in Fig. 5.

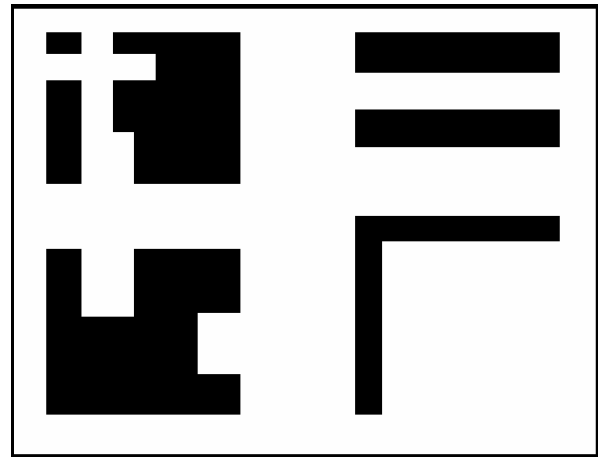

Fig. 5. The learning environment

This environment was chosen so that it contains various situations, which can occur during the robot motion.

As a testing environment was used the bit map of the laboratory environment as is depicted at the Fig. 6. These environments were scanning with ultrasound range finder. The model of the range finder and sensing of its data was implemented as follows: virtual environment with obstacles was represented as a bit map. The range finder scanned this environment by emitting beams (see Fig. 6). If the beam in competent direction collided with obstacles (the bit value 0 – black color) we calculate this distance $d_i$. The calculated and normalized distances $d_i$ was used as inputs to the neural networks.

Scanning of the environment was in 29 directions. This number was obtaining from many experiments and simulations. At these 29 directions obtained distances $d_i$ create input patterns to the PCA network.

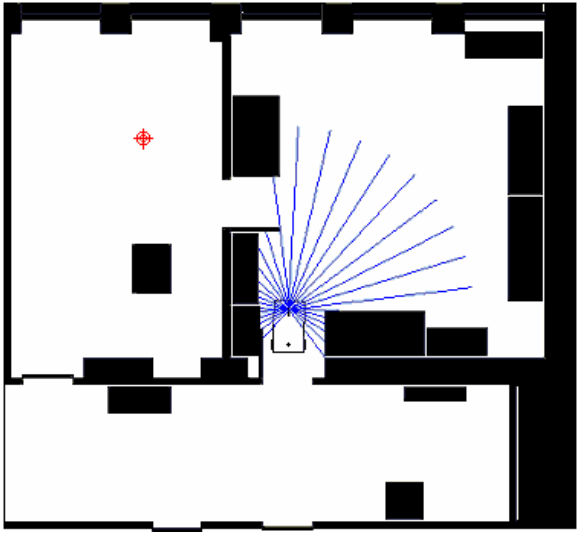

Fig. 6. The scanning range finder

The outputs from this network are free environment segments as are depicted on Fig. 3.

For the learning of the free segments was create the safe sensor map. It is the set of minimal distances from obstacles for actual state of the scanning scene using the really dimensions of the robot. At the robot motion (and scanning) in learning environment we compare values $d_i$ with the safe sensor map at the each step and we determine the free segments. The obtained information $d_i$ (29 values) and information about free segments $V_i$ (9 values) was in each step saved to the file. This file forms the training set of pattern for PCA network in second phase.

The second neural network - a feed-forward network - is typically trained with static back-propagation algorithm. For the learning of this supervised MLP network was using the combination status table. This table contains all potential combinations of the space segments $V_i$ and the goal segments $S_j$. The parameters $V_i$ and $S_j$ have two values in the table. If $V_i$ =1 then space segment $V_i$ not containing the obstacle meaning the motion in this segment is possible. If its value is 0 the segment is occupied and the motion within it is not possible. The value $S_j$=1 says that in this goal segment the goal coordinates take place and the value 0 signalises the absence of the goal coordinates. The supervisor attaches the required output (as the robot's reaction) to any combination of $V_i$ and $S_j$, and the network learns these situations. The robot reaction (azimuth) the supervisor chooses from free $V_i$ and $S_j$ so as the robot motion was realized in the safe direction to the goal. Our network had 18 input neurons (9 value $V_i$ and 9 value $S_j$), 20 hidden neurons and 9 output neurons (azimuth $O_j$).

Output $O_j$ of this second network is given to the control unit. It manages this information into the robot command for the robot motor control. In our case experimental robot distinguish five motion commands: forward, turn left, left, turn right, right. The commands turn left (right) means turning about 45° in left (right) direction.

The commands left (right) means turning about 90° in left (right) direction.

When we testing the functionality of proposed algorithm we find out the critical location in which the robot do not know continue in correctly direction. There was a door in the laboratory room or the narrow location in corridor. Therefore we add two neural networks (network H and network D) the multilayer perceptron type. The aim of the network H is recognize when the robot is situated in "hazard" – it stay in narrow location. When this situation is finding the network D execute the safe motion through this narrow location (see Fig. 7).

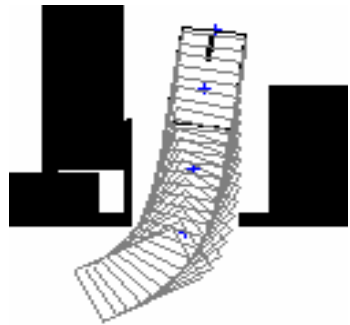

Fig 7. Motion by the door

The network H operates as a switch, which decides about the using the outputs from network MLP (described in section 4. 2.) or from network D for robot motion.

## 5. Simulation results

In our laboratory we have experimental mobile robot AURO, see Fig. 8. It is built up as a prism platform with three-wheeled configuration, which has a length of 850 mm, a width of 500 mm, and a height of 750 mm. It consists of single steerable drive wheel at the front and two passive rear wheels. Two stepper motors are used for driving and steering the front wheel. It has a capability of motion in longitudinal directions and rotation around the robot's reference point and it can reach a maximum speed of 0.1 m/sec. The drive wheel as well as the passive wheels is equipped with shaft encoders used for odometry measurement. For sensing the environment it has ultrasonic scanning range finder, rotating from 0° to 360°. and three tactile sensors on the bumper. The robot should travel from its initial position to a final desired position across a two-dimensional structured environment. The robot obtains range images by an ultrasound scanning range finder. The ranges for desired angular sector are obtained in N steps, covering up to 180° arcs in front of the robot, are measured by scanning every 200 ms by time-of-flight principle (Uher, L. & Kello, I. 1999). The ultrasound scanning range finder is disposed on the robot to get the distance measure $d_i$ in the vicinity of the robot where 20 cm < $d_i$ < 250 cm. The main navigation level computation is performed on a host PC via RS 232 communication. The robot manoeuvres are controlled by delivering information of rotation velocity and heading angle of the front wheel.

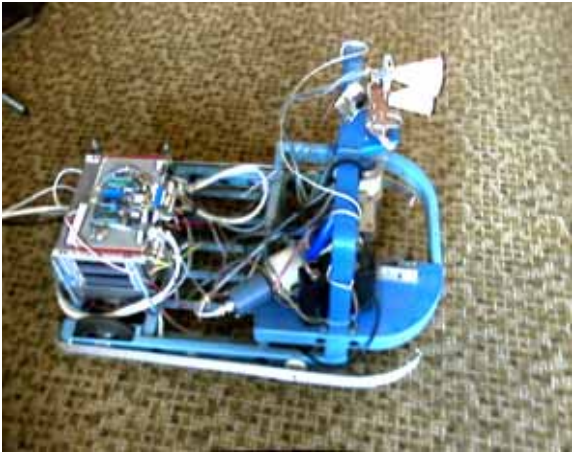

Fig. 8. Experimental mobile robot

We have implemented the algorithm described in the above sections in a path planner program Neuro written down in the language C++ on PC Intel Pentium 350 MHz. For the learning and testing of the network we used the programs from NeuroSolutions packages (Neurodimension 2000). The programs were interconnected with help of the dynamic data change (DDE) in order to enable using the data from program packages NeuroSolutions in the program Neuro.

Several examples were used to test our algorithm at first for a point robot. These first results were presented in (Janglova, 2000).

The functionality of the proposed algorithm was tested in a few type of the workspace. First tests were in the learning environment. Here the robot avoids to all obstacles and it executes each path from the giving start point to the goal point safely. Next testing examples were doing in the environment that was not use for learning of network, i.e. the unknown environment for the robot.

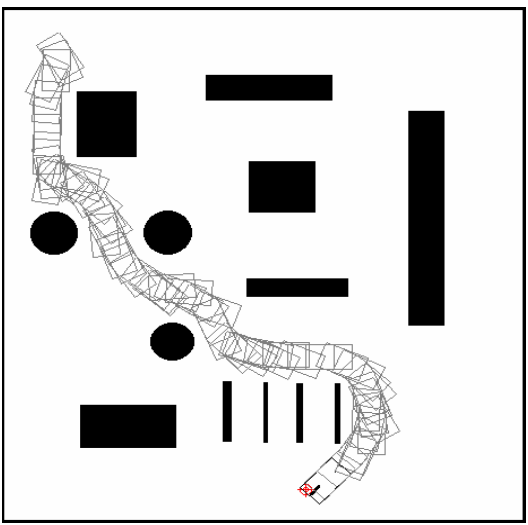

Fig. 9. Robot path in unknown environment

At the Fig. 9 is shown this situation – the execute path is collision-free.

Then we use this algorithm for simulation of the motion for our experimental mobile platform in the laboratory environment.

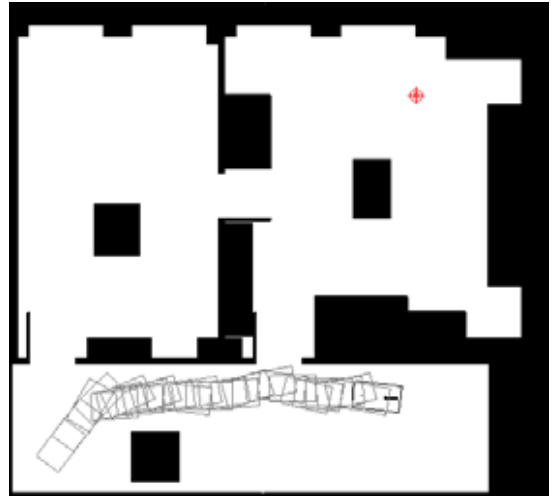

Fig. 10. Simulation of robot motion

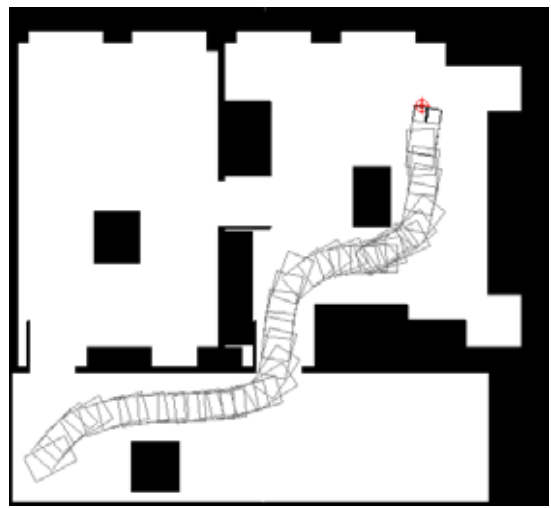

Fig. 11. Robot motion from corridor to the room

The obtained results are in the Fig. 10. and Fig. 11. In these figures are given the bit map of indoor environment and the paths, which was generated by the above designed algorithm.

The Fig. 10. shows simulation of the robot path when robot task had moving from the left corner of corridor to the goal marked by cross at the laboratory room. The robot does not reach the goal position – it was not pass through the door.

This same simulation example with using the adding neural network D and H is shown on the Fig. 11. It is seen that the robot path is collision-free and safe from the start to t he goal position. From the shown examples we conclude that this strategy is usable in general for motion of the robot in arbitrary environment.

## 6. Conclusion

The paper presents our first results that we obtained making use of the proposed path planning algorithm working with the neural network and sensor data. The simulation examples of the generation of the collision-free path for point robot and for two-dimensional robot show that designed strategy are acceptable for solution of this problem. We played the role of the supervisor to learn the robot to make it's way intelligently toward its target and to avoid obstacles.

In future we will implement this technique for safe motion of our experimental mobile vehicle in indoor conditions. We suppose to use this algorithm not only for the robot motion in known environment but for unknown one, as well. It is necessary to test different parameters in neural network with the aim of reaching the optimal time for finding the (shortest possible) safe path. As the robot collects environment data currently along its path it can avoid not only the static obstacles but also the dynamic ones. We feel that this technique will be suitable also for the motion of mobile devices in complex environment comprising also mobile obstacles.

ACKNOWLEDGEMENT

This work was supported VEGA MŠ SR a SAV grant No. 2/3129/23.

RoboCup
Portugal - 2004

LISBOA
PORTUGAL
27 June - 5 July 2004

RoboCupSoccer

RoboCupSoccer